\documentclass[conference]{IEEEtran}
\ifCLASSINFOpdf
  \usepackage[pdftex]{graphicx}
\else
  \usepackage[dvips]{graphicx}
\fi

\usepackage{graphicx}
\usepackage{subfigure}
\usepackage{times}
\usepackage{graphics}
\usepackage{amsmath, amsthm, amssymb, multirow, paralist, mathtools}
\usepackage{url}
\usepackage{algorithm,algorithmic}

\newtheorem{thm}{Theorem}

\def \bqs {\begin{eqnarray*}}
\def \eqs {\end{eqnarray*}}
\def \bq {\begin{eqnarray}}
\def \eq {\end{eqnarray}}

\def \x {\mathbf{x}}

\def \R {\mathbb{R}}

\def \prox {\mathrm{prox}}
\def \sign {\mathrm{sign}}

\def \Tr   {\mathrm{Tr}}

\begin{document}
%
\title{ Sparse Online Relative Similarity Learning}

\author{\IEEEauthorblockN{Dezhong Yao\IEEEauthorrefmark{1}, Peilin Zhao\IEEEauthorrefmark{2}, Chen Yu\IEEEauthorrefmark{1}, Hai Jin\IEEEauthorrefmark{1} and Bin Li\IEEEauthorrefmark{3}}
\IEEEauthorblockA{\IEEEauthorrefmark{1}Services Computing Technology and System Lab, Cluster and Grid Computing Lab\\
School of Computer Science and Technology, Huazhong University of Science and Technology, Wuhan, 430074, China\\}
\IEEEauthorblockA{\IEEEauthorrefmark{2}Data Analytics Department, Institute for Infocomm Research, A*STAR, 138632, Singapore\\}
\IEEEauthorblockA{\IEEEauthorrefmark{3}Economics and Management School, Wuhan University, Wuhan, 430072, China\\
Email: \{dyao, yuchen, hjin\}@hust.edu.cn, zhaop@i2r.a-star.edu.sg, binli.whu@whu.edu.cn}
}

\maketitle

\begin{abstract}
For many data mining and machine learning tasks, the quality of a similarity measure is the key for their performance. To automatically find a good similarity measure from datasets, metric learning and similarity learning are proposed and studied extensively. Metric learning will learn a Mahalanobis distance based on \emph{positive semi-definite} (PSD) matrix, to measure the distances between objectives, while similarity learning aims to directly learn a similarity function without PSD constraint so that it is more attractive. Most of the existing similarity learning algorithms are online similarity learning method, since online learning is more scalable than offline learning. However, most existing online similarity learning algorithms learn a full matrix with $d^2$ parameters, where $d$ is the dimension of the instances. This is clearly inefficient for high dimensional tasks due to its high memory and computational complexity. To solve this issue, we introduce several \emph{Sparse Online Relative Similarity} (SORS) learning algorithms, which learn a sparse model during the learning process, so that the memory and computational cost can be significantly reduced. We theoretically analyze the proposed algorithms, and evaluate them on some real-world high dimensional datasets. Encouraging empirical results demonstrate the advantages of our approach in terms of efficiency and efficacy.
\end{abstract}


%
\IEEEpeerreviewmaketitle

\section{Introduction}

In many applications, it is important to have a suitable similarity measure  between objectives, such as  text mining~\cite{DBLP:journals/ldvf/HothoNP05},  image retrieval~\cite{DBLP:conf/cvpr/HoiLLM06}. For example, when using $k$ \emph{nearest neighbor} ($k$-NN) to classify documents, the accuracy will significantly depend on the quality of the similarity measure. In addition, the performance of content based image retrieval also heavily relies on the similarity measure. To find a meaningful similarity measure, online metric learning has been extensively studied for years~\cite{DBLP:conf/nips/XingNJR02,GoldbergerRHS04,LMNNnipsWeinbergerBS05,icml_McFeeL10} due to the scalability of online learning~\cite{DBLP:conf/icml/ZhaoH10,DBLP:conf/icml/ZhaoHJY11,DBLP:journals/jmlr/ZhaoHJ11,HoiWZ14}, where a Mahalanobis distance function is learnt. Specifically, online metric learning approaches try to learn a  linear Mahalanobis distance ~\cite{de2000mahalanobis,LEGOjain2009online,OASISChechik2010,aaaiWuDZMH14}, satisfying a set of constraints. Generally the constraints assign larger similarity scores for pairs of similar instances than dissimilar instances. This is equivalent to learn a linear projection of the data to a feature space where constraints on the training set are better satisfied.  These metric learning algorithms have significantly improved the classification accuracy of $k$-NN and retrieval performance of content-based image retrieval on real-world datasets. As we know, Mahalanobis distance requires the model, i.e. the matrix, to be \emph{positive semi-definite} (PSD). So during the learning process, projections are needed to keep the model in the PSD cone. However, the projections need singular value decompositions with very high time complexity, which makes metric learning not scalable with respect to the dimension of instances.

To solve this issue, online similarity learning has been recently studied~\cite{OASISChechik2010,aaaiWuDZMH14,QamarG09,surveyBelletHS13,DBLP:conf/mm/WuHXZWM13,DBLP:journals/pami/XiaHJZ14}, which does not require its model to be PSD, so it is much more efficient than metric learning. In addition, the similarity metric can even be non-symmetric, so it can be applied to measure similarities of objects from two different feature spaces. Following this principle, similarity learning is studied for classification problems~\cite{QamarG09,IcmlBelletHS12}, and image retrieval tasks~\cite{OASISChechik2010,aaaiWuDZMH14}. Empirical studies show that those online similarity learning algorithms generally are better than or at least comparable with the metric learning algorithms on the classification tasks and image retrieval problems. At the same time, they are more efficient and scalable than metric learning algorithms, since the projection steps are avoided. As a result, online similarity learning is more suitable to large-scale data mining and machine learning tasks. However, these existing online similarity learning algorithms~\cite{OASISChechik2010,crammer2012adaptive} usually try to learn a full matrix during the learning process, which results in high memory and computational cost for high dimensional learning tasks.

Recently, to solve high dimensional metric/similarity learning problems, there are several off-line learning algorithms focused on learning sparse global linear metrics, e.g., SDML~\cite{SDMLQiTZCZ09}, SML~\cite{SMLYingHC09} and GSML~\cite{GSMLHuangYC11}. They are all based on Davis's work~\cite{ITMLdavis2007information} and try to learn a global sparse model by minimizing the Bregman divergence between the model and the identity matrix corresponding to the Euclidean distance, subject to a set of linear constraints. By minimizing Bregman divergence, the learned Mahlanobis matrix usually can be as close to the identity matrix as possible~\cite{SDMLQiTZCZ09}, so that the learnt Mahlanobis matrix can be very sparse, since  the identity matrix is rather sparse. However, these algorithms cannot handle online learning scenario, which is more realistic for many real-world data stream tasks. Motivated by the above observations, we propose a new \emph{Sparse Online Relative Similarity Learning} (\textbf{SORS}) scheme, where the similarity function is a sparse model learnt efficiently and scalably from the training data. Specifically, we propose two types of algorithms: SORS based on recent proximal online gradient descent; and AdaSORS based on state-of-the-art adaptive proximal online gradient descent. The scalability and efficiency of these advanced sparse online learning strategies make the proposed algorithms attractive to large-scale high dimensional real world applications. The contributions of our work are summarized as following: 1) We introduce a new Sparse Online Relative Similarity learning scheme that takes sparsity into consideration; 2) Based on the scheme, we propose several sparse online similarity learning algorithms; 3) We theoretically analyze the proposed algorithm; 4) We apply the proposed algorithms on a set of real-world datasets, where encouraging empirical results are achieved compared with some state-of-the-art similarity/metric learning algorithms.

The rest of this paper is organized as follows. The related work is reviewed in the next section. Section ~\ref{sec:alg} presents our proposed sparse online relative similarity learning approaches. Section ~\ref{sec:exp} conducts an extensive set of experiments by comparison with several state-of-the-art methods, and section ~\ref{sec:con} concludes our paper.

\section{Related Work}
\label{sec:relwork}

\vspace{-0.05in}
\subsection{Similarity/Metric Learning}
\vspace{-0.05in}
Online similarity learning is a group of efficient and scalable machine learning algorithms~\cite{HoiWZ14,aaaiWuDZMH14,MDMLKunapuliS12}. Most existing works in similarity learning rely on learning a Mahalanobis distance, which has been found to be a sufficiently powerful class of metrics that work on many real-world problems. POLA~\cite{POLAShalev-ShwartzSN04}, \emph{Pseudo-metric Online Learning Algorithm}, is the first online Mahalanobis distance learning approach which learns the distance metric $M\succeq 0$ as well as a threshold $b\geq1$. The authors provide a regret bound for this algorithm, under the assumption that the dataset can be separated by some metrics. LMNN ~\cite{LMNNnipsWeinbergerBS05} addresses the distance learning problem by exploring a large margin nearest neighbor classifier LEGO~\cite{LEGOjain2009online}, \emph{LogDet Exact Gradient online}, an improved version of POLA based on LogDet divergence regularization. It features tighter regret bounds, more efficient updates, and automatic semi-definiteness. RDML~\cite{RDMLjin2009regularized}, \emph{Regularized Distance Metric Learning}, is also similar to POLA but is more flexible. At each step $t$, instead of forcing the margin constraint to be satisfied, it performs a gradient descent step. Those methods focus on the symmetric distance: measure the distance between two images $\x_1$ and $\x_2$ by $(\x_1-\x_2)^\top M(\x_1-\x_2)$, where the matrix $M$ must be positive semi-definite.

The proposed online learning scheme is close to the recent work of scalable image similarity learning (OASIS)~\cite{OASISChechik2010,aaaiWuDZMH14}. This method learns bilinear similarity with a focus on large-scale problems. In this model, given two images $\x_1$ and $\x_2$, it measures similarity through $\x_1^\top M\x_2$. The metric $M$ is related to the (generalized) cosine similarity but does not include normalization nor PSD (symmetric positive semi-definite) constraint. In another part, unlike the Mahalanobis distance, it can define a similarity measure between instances of different dimensions. However, the learnt model $M$ is not sparse, which is unsuitable to large scale tasks for high dimension datasets.

\vspace{-0.05in}
\subsection{Sparse Online Learning}
\vspace{-0.05in}
The general online learning algorithms have solid performance on classification problems. However, for large-scale high-dimensional task, they learn a full feature classifier, which may cause high memory demand and heavy computational load. To deal with this limitation, the \emph{Sparse Online Learning}~\cite{LangfordLZ09,DuchiS09,DBLP:conf/icdm/WangWZMH14} has been studied recently. The goal of sparse online learning is to learn a sparse classifier, which only contains limited active features. In current \emph{Sparse Online Learning} studies, there are two types of solutions. One solution follows the general idea of subgradient descent with truncation, such as, FOBOS method~\cite{DuchiS09}, makes an improvement based on the \emph{Forward-Backward Splitting} method to solve the sparse online learning problem. There are two steps in the FOBOS method: (1) an unconstrained subgradient descent step, and (2) an instantaneous optimization step for a trade off between maintaining the result calculated in the first step and minimizing $\ell_{1}$ norm regularization. The optimization goal in the step 2 can be solved by adopting \emph{soft-thresholding} operations, which make truncation on the weight vectors. Under the guidance of this idea, the TG scheme~\cite{LangfordLZ09}, \emph{Truncated Gradient}, is proposed. When there are less than a predefined threshold $\theta$, the TG scheme truncates coefficients every $K$ steps. This demonstrates that truncation on each iteration is too aggressive as each step modifies the coefficients by only a small amount. The second solution of sparse online learning uses the dual averaging method~\cite{Nesterov09}, which studies the regularization structure in an online setting. RDA~\cite{Xiao10}, \emph{Regularized Dual Averaging}, is one representative study, which simultaneously achieves the optimal convergence rates for both convex and strongly convex loss. An extension work~\cite{LeeW12} of RDA algorithm is presented by using a more aggressive
truncation threshold and generates significantly more sparse solutions. However, all the above algorithms are designed for classification problems.

\section{Sparse Online Relative Similarity Learning}
In this section, we will review the problem setting,  propose our algorithm, and provide theoretical guarantees.
\label{sec:alg}

\vspace{-0.05in}
\subsection{Problem Formulation}
\vspace{-0.05in}
Following~\cite{OASISChechik2010}, we would study the problem of online similarity learning. Our goal is to learn a similarity function $S:\R^d\times\R^d\rightarrow \R$ based on a sequence of training triplets $\{(\x_t,\x_t^+, \x_t^-)\in \R^d \times \R^d \times \R^d| t\in[T]\}$  with $[T]=\{1, \ldots, T\}$, where the relevance between $\x_t$ and $\x_t^+$ is greater than that between $\x_t$ and $\x_t^-$. Specifically, we would like to learn a similarity function $S(\x,\x')$ that assigns higher similarity scores to more relevant instances, i.e.,
\begin{eqnarray*}\label{eqn:similarity function}
S(\x_t,\x_t^+) > S(\x_t, \x_t^-), \forall t\in[T]
\end{eqnarray*}

For example, compared with an image $\x_t^-$, $\x_t^+$ is a more similar image with $\x_t$.

For the similarity function, we  adopt a parametric similarity function that has a bi-linear form,
\begin{eqnarray*}\label{eqn:bilinear similarity function}
S_M(\x,\x') = \x^\top M \x'
\end{eqnarray*}
where $M\in\R^{d\times d}$. In order to learn the optimal parameter $M$, we introduce a loss function $\ell(M;(\x_t,\x_t^+,\x_t^-))$ that measures its performance on the $t$-th triplet. One popular loss function is the well-known hinge loss
\begin{eqnarray*}
\ell(M;(\x_t,\x_t^+,\x_t^-))=[1-S_M(\x_t,\x_t^+)+S_M(\x_t,\x_t^-)]_+
\end{eqnarray*}
where $[\cdot]_+=\max(0, \cdot)$. The above loss measures how much the violation of the desired constraint $S_M(\x_t,\x_t^+) \geq S_M(\x_t,\x_t^-)$ is by the similarity function defined by $M$.

Under these settings, an online learning algorithm updates its model in rounds. At each round a new model is estimated, based on the previous one and the current triplet. We denote the models estimate after the $t-1$-th round by $M_t$. A typical online algorithm starts with some initial models, e.g., $M_1=0$. For each round $t$, a triplet $(\x_t,\x_t^+,\x_t^-)$ will be presented to the algorithm. The algorithm can update the model from $M_t$ to $M_{t+1}$, based on the current triplet. The goal of online similarity learning is to minimize the regret of the algorithm defined as
\begin{eqnarray}\label{eqn:regret}
R_T = \sum^T_{t=1}\ell_t(M_t)-\min_M\sum^T_{t=1}\ell_t(M)
\end{eqnarray}
where $\ell_t(M)=\ell(M;(\x_t,\x_t^+,\x_t^-))$. If the regret is sublinear, i.e., $R_T\le o(T)$, it is easy to observe
\begin{eqnarray}
 \sum^T_{t=1}\ell_t(\bar{M}_T)\le \min_M\sum^T_{t=1}\ell_t(M) + o(T)
\end{eqnarray}
where $\bar{M}_T=\frac{1}{T}\sum^T_{t=1}M_t$, so  $\bar{M}_T$ is a good estimate of $\arg\min_M\sum^T_{t=1}\ell_t(M)$. In practice, The final outputted model $M_T$ usually will be used for testing. Please note the online triplets  are generally not provided in advance.

To solve this problem, some authors proposed an online learning scheme, OASIS in~\cite{OASISChechik2010}, by applying the online \emph{Passive Aggressive} (PA) learning strategy~\cite{DBLP:journals/jmlr/CrammerDKSS06}, i.e.,
\begin{eqnarray*}
M_{t+1} = \arg\min_{M}\frac{1}{2}\|M-M_t\|^2_F +C \ell_t(M)
\end{eqnarray*}
where $C>0$ is a trade-off parameter for the regularization term and loss. This algorithm enjoys the following closed-form solution
\begin{eqnarray*}
M_{t+1} = M_t + \tau_t \x_t(\x_t^+-\x_t^-)^\top
\end{eqnarray*}
where $\tau_t=\min(C, \frac{\ell_t(M_t)}{\|\x_t(\x_t^+-\x_t^-)^\top\|^2_F})$, and $\|\cdot\|_F$ is Frobenius norm. This algorithm is only provided with a mistake bound by the authors. However, a careful analysis should give it a regret bound of order $O(\sqrt{T})$.

 Another possible method to solve this problem is based on the \emph{Online Gradient Descent} (OGD) method~\cite{zinkevich2003online}, i.e.,
\begin{eqnarray*}
M_{t+1} = M_t - \eta_t G_t
\end{eqnarray*}
where $G_t=\partial_M  \ell_t(M_t)$. The gradient $G_t$ can be computed as follows
\begin{displaymath}
G_t = \left\{ \begin{array}{ll}
- \x_t(\x_t^+-\x_t^-)^\top & \textrm{if $\ell_t(M_t)>0$}\\
0 & \textrm{else}
\end{array} \right.
\end{displaymath}

This algorithm generally has a regret bound of $O(\sqrt{T})$. However these two algorithms both learn a dense matrix $M$, which may result in high testing time complexity and memory cost. To solve this issue,  we will study sparse online learning techniques to tackle this problem in this paper.



\vspace{-0.05in}
\subsection{Sparse Online Relative Similarity Learning (SORS)}
\vspace{-0.05in}
To make the model sparse, we can add sparse regularization function into the online objective function, i.e.,
\begin{eqnarray}\label{eqn:online-objective}
&&\hspace{-0.4in}M_{t+1}= \arg\min_{M}\Big[\ell_t(M) \hspace{-0.02in}+\lambda r(M) \hspace{-0.02in}+\frac{\|M\hspace{-0.02in}-\hspace{-0.02in}M_t\|^2_F}{2\eta_t}\Big]
\end{eqnarray}
where $r(M)$ is typically set as
\bqs
r_1(M)=\|M\|_1=\sum_{ij}|M_{ij}|
\eqs

This objective function has three terms:
\begin{itemize}
\item the first one is to minimize the loss of new similarity matrix on the current triplet;
\item the second term is to make the new similarity matrix sparse, where $\lambda$ is a sparsity parameter. If $\lambda$ is larger, the model will be sparser.
\item the final term is used to keep the new similarity matrix close to the current one.
\end{itemize}

The reason to use $\|M\|_1$ is that $\|M\|_1$ is a convex upper bound of $\|M\|_0$, which can make the optimal solution sparse, i.e., most of $M_{ij}$ will be zeros. In this way, the computation of $S(\x,\x')$ will be significantly reduced, since $S(\x,\x')=\sum_{ij}M_{ij}\x_i\x'_j$. However, this regularization treats the diagonal elements of $M$ equally with those off diagonal elements, it is clearly insufficient,  since diagonal elements should generally be dense while  those off-diagonal elements should be sparse. The reason is that the diagonal element represents the self-correlation of a feature while the off-diagonal elements represent correlation between features. Self-correlations are always larger while inter correlations are generally sparse in high dimension space. To solve this issue, an alternative is to adopt the off-diagonal $L_1$ norm
\bqs
r_2(M)=\|M\|_{1,\text{off}}=\sum_{i\not=j}|M_{ij}|
\eqs
to pursue a sparse solution while keep the diagonal elements dense.

Given the regularization function, we would solve the online objective function Eq.~\eqref{eqn:online-objective}. However this optimization usually does not have a closed-form solution, which will incur additional computational cost. To tackle it, we propose to linearize the loss function $\ell_t(M)$, to get the following new online objective function
\begin{eqnarray}\label{eqn:goal}
&&\hspace{-0.4in} M_{t+1}\hspace{-0.03in}\nonumber\\
&&\hspace{-0.4in}=\arg\min_{M}\hspace{-0.03in}\Big[\ell(M_t)+\hspace{-0.03in}\langle M-M_t, G_t\rangle  \hspace{-0.03in}+\hspace{-0.03in}\lambda r(M) \hspace{-0.03in}+\hspace{-0.03in}\frac{\|M-M_t\|^2_F}{2\eta_t}\hspace{-0.02in}\Big]\nonumber\\
&&\hspace{-0.4in}=\arg\min_{M}\hspace{-0.03in}\Big[\hspace{-0.03in}\langle M, G_t\rangle  \hspace{-0.03in}+\hspace{-0.03in}\lambda r(M) \hspace{-0.03in}+\hspace{-0.03in}\frac{\|M-M_t\|^2_F}{2\eta_t}\hspace{-0.02in}\Big]
\end{eqnarray}
where $\langle M, G_t\rangle=\Tr(M^\top G_t)$ and $G_t=\partial_M\ell_t(M_t)$.

It is easy to verify that this objective will produce the $t+1$-th iterate as
\bq\label{eqn:SORSproxfunc}
M_{t+1} = \prox_{\eta_t\lambda r}(M_t-\eta_t G_t)
\eq
where $\prox_h(M)=\arg\min_N\Big(h(N)+\frac{1}{2}\|N-M\|^2_F\Big).$
Luckily, when $r_1(M)=\|M\|_1$, we have
\bq\label{eqn:SORSrank1}
\prox_{\eta_t\lambda r_1}(M)=\sign(M)\odot[|M|-\eta_t\lambda]_+
\eq
and when $r_2(M)=\|M\|_{1,\text{off}}$, we have
\bq\label{eqn:SORSrank2}
\prox_{\eta_t\lambda r_2}(M)=\sign(M)\odot[|M| - \eta_t\lambda (1-I)]_+
\eq
where $I$ is the identity matrix. $\odot$ is element-wise product for two matrices. These updates can be computed efficiently.

Finally, we can summarize the proposed  Sparse Online Relative Similarity Learning algorithm in Algorithm~\ref{alg:SORS}.

\begin{algorithm}[htpb]
\caption{\textbf{S}parse \textbf{O}nline \textbf{R}elative \textbf{S}imilarity Learning {\bf(SORS)}} \label{alg:SORS}
\begin{algorithmic}[1]
\STATE {\bf Input}: Regularization parameter $\lambda>0$, learning rates $\eta_t>0$, and regularizer $r(\cdot)$.
\STATE {\bf Initialize}:  $M_1=I$.
\FOR{$t=1,\ldots,T$}
\STATE Receive $(\x_t,\x_t^+,\x_t^-)$;
\STATE Compute $\ell(M_t;(\x_t,\x_t^+,\x_t^-))$;
\STATE $G_t=\partial_M\ell(M_t;(\x_t,\x_t^+,\x_t^-))$;
\STATE $M_{t+1} = \prox_{\eta_t\lambda r}( M_{t}-\eta_t G_t)$;
\ENDFOR
\STATE {\bf Output}: $M_{T+1}$
\end{algorithmic}
\end{algorithm}

\vspace{-0.05in}
\subsection{Adaptive Sparse Online Relative Similarity Learning (AdaSORS)}
\vspace{-0.05in}
To improve the learning performance of relative similarity learning, a second order similarity learning algorithm AROMA is proposed~\cite{DBLP:conf/icml/CrammerC12}. AROMA is based on the well-known Confidence Weighted learning strategy~\cite{DBLP:conf/icml/HoiWZ12}, which does not only use the first order information, i.e., weighted mean of examples, but also the second-order information, i.e., the covariance matrix for all the features, to update the model. Theoretical results and empirical comparisons have shown that AROMA can significantly outperform OASIS in terms of information retrieval performance. However, the model produced by AROMA is still dense.

In this subsection, we would like to adopt the second order information of the data stream to further improve the learning efficiency and efficacy of sparse online relative similarity learning algorithm. Different from AROMA, we will utilize the idea of AdaGrad~\cite{jmlrDuchiHS11,DBLP:conf/aaai/DingZHO15} to incorporate the second order information through adaptive regularization. Specifically, we will maintain a matrix $\Sigma_t= \delta  + H_t$, where $\delta$ is a smooth parameter to keep each element of $\Sigma_t$ invertible and $H_t$ is similar with correlation matrix between features. Specifically, $H_1$ will be initialized as zero matrix, and at the $t$-th iteration, it will be updated using the following rule:
\bqs
H_{t,ij} = \sqrt{H_{t-1,ij}^2+G_{t,ij}^2},\forall i,j\in[d]
\eqs

Given the matrix $\Sigma_t$, we can introduce a norm induced by $\Sigma_t$, which is defined as follow:
\bqs
\|M\|^2_{\Sigma_t} = \sum_{ij} \Sigma_{t,ij} M_{ij}^2
\eqs

Given this norm, we can follow the similar idea of AdaGrad to replace the final term of the  first-order objective function of SORS with this new norm, to get the following updating strategy
\begin{eqnarray}\label{eqn:so-goal}
&&\hspace{-0.4in} M_{t+1}\hspace{-0.03in}=\arg\min_{M}\hspace{-0.03in}\Big[\hspace{-0.03in}\langle M, G_t\rangle  \hspace{-0.03in}+\hspace{-0.03in}\lambda r(M) \hspace{-0.03in}+\hspace{-0.03in}\frac{\|M-M_t\|^2_{\Sigma_t}}{2\eta_t}\hspace{-0.02in}\Big]
\end{eqnarray}
which is considered as the objective of Adaptive Sparse Online Relative Similarity learning algorithm. The unique difference between this objective and SORS is the final proximity term.

If we define a proximal operator as
\bqs
\prox_{h}^{\Sigma_t}(M)=\arg\min_N(h(N)+\frac{1}{2}\|N-M\|^2_{\Sigma_t})
\eqs
then, it is easy to see the solution for AdaSORS is
\bq\label{eqn:AdaSORSproxfunc}
M_{t+1}=\prox_{\eta_t\lambda r}^{\Sigma_t}(M_t-\eta_t G_t./\Sigma_t)
\eq
where $./$ is element-wise division.

When $r$ is set as $r_1=\|M\|_1$, the corresponding proximal operator is
\bq\label{eqn:AdaSORSrank1}
\prox_{\eta_t\lambda r_1}^{\Sigma_t}(M)=\sign(M)\odot[|M|-\lambda\eta_t./\Sigma_t]_+
\eq
and when $r$ is set as $r_2=\|M\|_{1,off}$, the corresponding proximal operator is
\bq\label{eqn:AdaSORSrank2}
\prox_{\eta_t\lambda r_2}^{\Sigma_t}(M)=\sign(M)\odot[|M|-\lambda\eta_t(1-I)./\Sigma_t]_+
\eq
whose time complexities are $O(d^2)$ so that they can be conducted efficiently.

Finally, we can summarize the proposed Adaptive Sparse Online Relative Similarity Learning algorithm in Algorithm~\ref{alg:AdaSORS}.
\begin{algorithm}[!htpb]
\caption{\textbf{Ada}ptive \textbf{S}parse \textbf{O}nline \textbf{R}elative \textbf{S}imilarity Learning {\bf(AdaSORS)}} \label{alg:AdaSORS}
\begin{algorithmic}[1]
\STATE {\bf Input}: Regularization parameter $\lambda>0$, learning rates $\eta_t>0$, smooth parameter $\delta$, and regularizer $r(\cdot)$.
\STATE {\bf Initialize}:  $M_1=I$.
\FOR{$t=1,\ldots,T$}
\STATE Receive $(\x_t,\x_t^+,\x_t^-)$;
\STATE Compute $\ell(M_t;(\x_t,\x_t^+,\x_t^-))$;
\STATE $G_t=\partial_M\ell(M_t;(\x_t,\x_t^+,\x_t^-))$;
\STATE $H_{t,ij} = \sqrt{H_{t-1,ij}^2+G_{t,ij}^2}$, $\forall i,j\in[d]$;
\STATE $\Sigma_t = \delta + H_t$;
\STATE $M_{t+1} = \prox_{\eta_t\lambda r}^{\Sigma_t}( M_{t}-\eta_t G_t./\Sigma_t)$;
\ENDFOR
\STATE {\bf Output}: $M_{T+1}$
\end{algorithmic}
\end{algorithm}

\vspace{-0.05in}
\subsection{Time Complexity Analysis}
\vspace{-0.05in}
The proposed algorithms' computational complexities depend on the number of features ($d$) of an instance. During each online learning step, element-wise operations are involved between $M_t\in\R^{d\times d}$ and $G_t\in\R^{d\times d}$, so the time complexity of SORS is $O(d^2)$ for each learning iteration. AdaSORS needs to deal with matrix $\Sigma_t$ and $H_t$, it consumes double time than SORS: $O(2d^2)$. If the instances are sparse, then generally we can compute and update the matrices $G$, $M$, $\Sigma$, and $H$ more efficiently.


\vspace{-0.05in}
\subsection{Theoretical Analysis}
\vspace{-0.05in}
\label{analysis}
In this subsection, we will provide theoretical upper bounds for the regret of the proposed algorithms defined as follow,
\begin{eqnarray*}
R_T=\sum^T_{t=1}[\ell_t(M_t)+\lambda r(M_t)]-  \sum^T_{t=1}[\ell_t(M_*)+\lambda r(M_*)]
\end{eqnarray*}
where $M_*=\arg\min_M \sum^T_{t=1}[\ell_t(M)+\lambda r(M)]$. This regret is different from the definition in the Eq.~\eqref{eqn:regret}, since this regret also considers the sparsity regularization term. The bound on the regret implies the gap between the cumulative loss of online learning algorithm and the cumulative loss of the best model which can be chosen in hindsight.

Now, we will analyze the theoretical performance of the proposed algorithms for the case that $\ell_t$ is Lipschitz continuous e.g., $\ell_t(M)=[1-\x_t^\top M\x_t^+ +\x_t^\top M\x_t^-]_+$.
\begin{thm}
Let $\{(\x_t,\x_t^+,\x_t^-)\ |\ t=1,\ldots,T \}$ be a sequence of triplets, where $\x_t,\x_t^+,\x_t^-\in\R^d$. Suppose $\ell_t(M)$ is  convex over $\R^{d\times d}$ and $L$-Lipschitz for any $t\in[T]$, if the SORS algorithm is run on this sequence of triplets with $\eta_t=\eta$, then we have the following regret bound,
\begin{small}
\bqs
R_T\le \frac{1}{2\eta}\|M_*\|_F^2 + \frac{\eta}{2} L^2T
\eqs
\end{small}
Furthermore, if we set $\eta=\|M_*\|/(L\sqrt{T})$, then for all $T>0$, we have
\begin{small}
\bqs
R_T\le\|M_*\|_F L\sqrt{T}
\eqs
\end{small}
\end{thm}

{\bf Remark}: This theorem implies the SORS algorithms will suffer a regret of order $O(\sqrt{T})$ and an average of regret of order $O(1/\sqrt{T})$. If $T$ approaches infinity, the average regret vanishes.

\begin{thm}
Let $\{(\x_t,\x_t^+,\x_t^-)\ |\ t=1,\ldots,T \}$ be a sequence of triplets, where $\x_t,\x_t^+,\x_t^-\in\R^d$. Suppose $\ell_t(M)$ is convex over $\R^{d\times d}$ and the gradient of $\ell_t$ with respect $M_t$ is $\nabla_M\ell_t(M_t)=G_t$, if the AdaSORS algorithm is run on this sequence of triplets, then we have the following regret bound
\bqs
R_T\le\frac{1}{2\eta}D_M^2\sum^d_{i=1}\sum^d_{j=1}\|G_{1:t,ij}\|_2+\eta\sum^d_{i=1}\sum^d_{j=1}\|G_{1:t,ij}\|_2
\eqs
where $D_M=\max_t\|M_*-M_t\|_\infty$ and $\|G_{1:t,ij}\|_2=\sqrt{\sum^T_{t=1}G_{t,ij}^2}$.

Furthermore, if $\eta$ is set as $D_\infty/\sqrt{2}$, we have
\bqs
R_T\le\sqrt{2}D_\infty\sum^d_{i=1}\sum^d_{j=1}\|G_{1:t,ij}\|_2
\eqs
\end{thm}
{\bf Remark:} According to~\cite{jmlrDuchiHS11}, this regret is in the same order of the regret by choosing the best fixed $\Sigma$ for the Eq.~\eqref{eqn:so-goal}.

\section{Experiments}
\label{sec:exp}
In this section, we evaluate the proposed approaches on six public high dimensional benchmark datasets to examine their effectiveness.

\vspace{-0.05in}
\subsection{Compared Algorithms}
\vspace{-0.05in}
We compare the following approaches:
\begin{itemize}
  \item \textbf{Euclidean}: The baseline measurement method using the standard Euclidean distance in feature space.
  \item \textbf{LMNN}: \emph{Largest Margin Nearest Neighbor} method proposed by~\cite{LMNNnipsWeinbergerBS05}. 
  \item \textbf{ITML}: \emph{Information Theoretic Metric Learning} proposed by~\cite{ITMLdavis2007information}. 
  \item \textbf{LEGO}: \emph{LogDet Exact Gradient Online} Learning proposed by~\cite{LEGOjain2009online}. 
  \item \textbf{RDML}: \emph{Regularized Distance Metric Learning} proposed by~\cite{RDMLjin2009regularized}. 
  \item \textbf{AROMA}: \emph{Adaptive Regularization for Weight Matrices} proposed by~\cite{crammer2012adaptive}.
  \item \textbf{OASIS}: It learns a bilinear similarity based on online Passive Aggressive algorithm using triplet instances~\cite{OASISChechik2010}.
  \item \textbf{SORS}: The algorithm described above in Algorithm~\ref{alg:SORS}. \textbf{SORS-I} is the algorithm adopted the  Eq.~\eqref{eqn:SORSrank1} as $\prox$ function in Eq.~\eqref{eqn:SORSproxfunc}. \textbf{SORS-II} is the one adopted the Eq.~\eqref{eqn:SORSrank2}.
  \item \textbf{AdaSORS}: The algorithm described above in Algorithm~\ref{alg:AdaSORS}. \textbf{AdaSORS-I} is the algorithm adopted the  Eq.~\eqref{eqn:AdaSORSrank1} as $\prox$ function in Eq.~\eqref{eqn:AdaSORSproxfunc}. \textbf{AdaSORS-II} is the one adopted the Eq.~\eqref{eqn:AdaSORSrank2}.

\end{itemize}

\vspace{-0.05in}
\subsection{Experimental Datasets and Setup}
\vspace{-0.05in}
To examine the performance, we test all the algorithms on six publicly available high dimensional datasets ``Caltech256''\footnote{\url{http://www.vision.caltech.edu/Image_Datasets/Caltech256}}, ``BBC''\footnote{\url{http://mlg.ucd.ie/datasets/bbc.html}}, ``Gisette''\footnote{\url{http://archive.ics.uci.edu/ml/datasets.html}}, ``Protein'', ``RCV1'', and ``Sector'' from LIBSVM\footnote{\url{http://www.csie.ntu.edu.tw/~cjlin/libsvmtools/datasets}}, as shown in Table~\ref{tb:hd-datasets}. The sparsity in Table~\ref{tb:hd-datasets} means the percentage of empty attribute values.

\begin{table}[!ht]
\vspace{-0.0in}
\centering
\caption{Statistics of Experimental Datasets.\label{tb:hd-datasets}}
\vspace{-0.0in}
\begin{tabular}{|c|c|c|r|r|c|}
\hline 
Data Set & Source & Class & \# Feature & \# Size & Sparsity (\%) \\ \hline \hline
Protein & LIBSVM & 3 & 357 & 17,766 & 71.00 \\
Caltech256  & Caltech & 257 & 1,000 & 29,461 & 96.75 \\
Gisette & UCI & 2 & 5,000 & 6,000 & 00.91 \\
BBC  & UCD & 5 & 9,636 & 2,225 & 98.66\\
RCV1 & LIBSVM & 53 & 47,236 & 8,967 & 99.86 \\
Sector & LIBSVM & 105 & 55,197 &  9,619 & 99.70 \\
\hline
\end{tabular}
\vspace{-0.1in}
\end{table}

Prediction for the structure and function of proteins plays an important role in bioinformatics. Protein dataset is used to predict the local conformation of the polypeptide chain~\cite{bibYang04}. This dataset comprises 17,766 protein entries with 3 categories. Each protein entry is represented by 357 features.

Caltech256 is a dataset for image retrieval tasks, which consists of 30,607 images assigned to 257 categories, where each image is represented by 1,000 features. We use the images from the first 10 and 50 classes to construct experimental datasets, which are denoted as ``Caltech10'' and ``Caltech50''.

Gisette is a dataset used for a handwritten recognition problem, which is to separate the confusable digits handwritten number '4' and '9'~\cite{confnipsGuyonGBD04}. The Gisette dataset is released from NIPS 2003 Feature Selection Challenge, which contains 6,000 entries. Each entry is represented by 5,000 features. From Table~\ref{tb:hd-datasets}, we can see that the entry in Gisette dataset is dense.

BBC news article dataset is gathered from BBC news website, which corresponds to stories in five topical areas from 2004-2005~\cite{icmlGreeneC06}. The dataset consists of 2,225 documents in five areas: business, entertainment, politics, sport and tech. Each document is represented by 9,636 features.

The Reuters RCV1 is a text classification dataset~\cite{jmlrLewisYRL04}. This dataset contains documents originally written in five different languages over a common set of 6 categories. Each RCV1 entry is represented by 47,236 features. We use the first 10 categories as our study dataset.

\begin{table*}[!htpb]
\caption{Evaluation of the Learning Algorithms on the Data Sets.}\label{tab:performance}
\renewcommand*\arraystretch{1.0}
\begin{center}
\begin{small}
\begin{tabular}{|l|c|c|c|c|c|c|c|}        \hline
\multirow{1}{*}{ Algorithm}& \multicolumn{3}{c|}{  Protein (\#Class=3, \#Iterations=$10^5$)} & & \multicolumn{3}{c|}{  Caltech50 (\#Class=50, \#Iterations=$10^5$) }\\
\cline{2-4} \cline{6-8}
 &   MAP (\%)         &   Sparsity (\%) & ~Time (s)~ &&    MAP (\%)         &   Sparsity (\%) & ~Time (s)~
\\\hline\hline
Eucl.	&	36.93$\pm$0.01	&	99.9$\pm$0.00	&	0	&	&	10.05$\pm$0.21	&	99.9$\pm$0.00	&	0	\\\hline
LMNN	&	45.91$\pm$0.01	&	0.56$\pm$0.01	&	78088.84$\pm$219.14	&	&	8.07$\pm$1.10 &	5.91$\pm$0.01	&	106884.06$\pm$894.25	\\
ITML	&	37.13$\pm$0.01	&	0.56$\pm$0.01	&	50.00$\pm$0.01	&	&	10.29$\pm$0.20	&	6.83$\pm$0.57	&	1124.88$\pm$26.01	\\
LEGO	&	37.44$\pm$0.01	&	0.56$\pm$0.01	&	60.00$\pm$0.01	&	&	10.15$\pm$0.01	&	5.52$\pm$0.01	&	531.94$\pm$4.62	\\
RDML	&	38.96$\pm$0.01	&	0.58$\pm$0.01	&	108.90$\pm$8.20	&	&	10.10$\pm$0.22	&	7.50$\pm$0.10	&	623.78$\pm$12.13	\\
OASIS	&	43.91$\pm$0.16	&	0.56$\pm$0.01	&	86.51$\pm$15.18	&	&	12.93$\pm$0.20	&	5.94$\pm$0.10	&	673.43$\pm$89.74	\\
AROMA	&	44.58$\pm$0.22	&	0.56$\pm$0.01	&	822.93$\pm$71.38	&	&	13.39$\pm$0.10	&	5.93$\pm$0.12	&	1294.66$\pm$147.32	\\
SORS-I	&	45.66$\pm$0.16	&	1.02$\pm$0.04	&	217.83$\pm$0.54	&	&	13.28$\pm$0.14	&	13.82$\pm$0.14	&	895.79$\pm$27.87	\\
SORS-II	&	45.69$\pm$0.06	&	0.81$\pm$0.03	&	239.62$\pm$1.18	&	&	13.29$\pm$0.18	&	13.83$\pm$0.18	&	958.97$\pm$20.19	\\
AdaSORS-I	&	47.02$\pm$0.41	&	0.97$\pm$0.05	&	400.11$\pm$9.21	&	&	13.82$\pm$0.14	&	14.10$\pm$1.21	&	2552.17$\pm$169.52	\\
AdaSORS-II	&	46.45$\pm$0.39	&	0.99$\pm$0.04	&	423.27$\pm$10.91	&	&	13.83$\pm$0.15	&	14.10$\pm$1.26	&	2993.64$\pm$282.13	\\
\hline\hline
\multirow{1}{*}{ Algorithm}& \multicolumn{3}{c|}{ Gisette (\#Class=2, \#Iterations=$10^5$)} & & \multicolumn{3}{c|}{  BBC (\#Class=5, \#Iterations=$10^5$) }\\
\cline{2-4} \cline{6-8}
 &   MAP (\%)         &   Sparsity (\%) & ~Time (s)~ &&    MAP (\%)         &   Sparsity (\%) & ~Time (s)~
\\\hline\hline
Eucl.	&	62.05$\pm$0.01	&	99.9$\pm$0.00	&	0		&	&	32.91$\pm$0.01	&	99.9$\pm$0.00	&	0	\\\hline
LMNN	&	87.01$\pm$0.24	&	1.75$\pm$0.01	&	78994.68$\pm$141.21		&	&NA	&NA &	NA				\\
ITML	&	62.09$\pm$0.32	&	28.64$\pm$0.01	&	750.82$\pm$23.12		&	&	31.78$\pm$0.47	&	10.86$\pm$1.43	&	109687.28$\pm$633.54	\\
LEGO	&	78.38$\pm$0.36	&	6.45$\pm$0.01	&	5410.54$\pm$30.94	&	&	30.29$\pm$0.01	&	3.44$\pm$0.26	&	31827.25$\pm$59.46	\\
RDML	&	82.49$\pm$0.32	&	20.73$\pm$0.01	&	2090.51$\pm$7.88		&	&	70.79$\pm$2.35	&	83.75$\pm$0.01	&	8189.38$\pm$188.18	\\
OASIS	&	83.49$\pm$0.22	&	4.72$\pm$0.01	&	2316.77$\pm$469.77		&	&	79.36$\pm$0.41	&	61.62$\pm$0.24	&	8881.26$\pm$475.22	\\
AROMA	&	83.57$\pm$1.82	&	3.89$\pm$0.18	&	26097.59$\pm$27.40		&	&	79.68$\pm$0.43	&	60.83$\pm$1.23	&	9723.02$\pm$381.83	\\
SORS-I	&	88.89$\pm$2.49	&	4.01$\pm$1.62	&	16839.32$\pm$101.85		&	&	91.03$\pm$1.12	&	78.67$\pm$3.56	&	90370.71$\pm$519.35	\\
SORS-II	&	89.89$\pm$1.22	&	4.08$\pm$1.61	&	22578.96$\pm$1028.02		&	&	92.20$\pm$1.28	&	78.41$\pm$3.21	&	135376.85$\pm$678.71	\\
AdaSORS-I	&	90.92$\pm$0.71	&	4.40$\pm$1.10	&	31256.19$\pm$1206.14		&	&	92.43$\pm$0.62	&	82.15$\pm$2.68	&	303718.57$\pm$643.43	\\
AdaSORS-II	&	90.99$\pm$0.53	&	4.20$\pm$1.12	&	35471.63$\pm$848.63		&	&	94.09$\pm$0.61	&	81.24$\pm$2.35	&	364976.73$\pm$1485.68	\\
\hline\hline
\multirow{1}{*}{ Algorithm}& \multicolumn{3}{c|}{ RCV1 (\#Class=10, \#Iterations=$10^5$) } & & \multicolumn{3}{c|}{ Sector (\#Class=105, \#Iterations=$10^5$) }\\
\cline{2-4} \cline{6-8}
 &   MAP (\%)         &   Sparsity (\%) & ~Time (s)~ &&    MAP (\%)         &   Sparsity (\%) & ~Time (s)~
\\\hline\hline
Eucl.	&	58.28$\pm$0.01	&	99.9$\pm$0.00	&	0	&	&	19.40$\pm$0.01	&	1$\pm$0.00	&	0	\\\hline
OASIS	&	92.37$\pm$0.01	&	97.14$\pm$0.01	&	30222.26$\pm$792.28	&	&	65.18$\pm$0.01	&	86.98$\pm$0.01	&	196561.51$\pm$1862.72	\\
AROMA	&	92.54$\pm$0.01	&	97.11$\pm$0.01	&	40407.61$\pm$149.85	&	&	65.47$\pm$0.01	&	87.91$\pm$0.01	&	275107.98$\pm$4295.58	\\
SORS-I	&	92.62$\pm$0.01	&	99.84$\pm$0.01	&	71473.62$\pm$1370.12	&	&	66.22$\pm$1.02	&	98.12$\pm$0.01	&	311849.83$\pm$3976.32	\\
SORS-II	&	92.62$\pm$0.01	&	99.84$\pm$0.01	&	100543.28$\pm$4981.78	&	&	66.89$\pm$1.02	&	98.12$\pm$0.01	&	381573.25$\pm$10874.97	\\
AdaSORS-I	&	92.82$\pm$0.01	&	99.64$\pm$0.01	&	148260.37$\pm$3411.35 &	&	67.18$\pm$0.01	&	98.20$\pm$0.01	&	632147.31$\pm$21041.49	\\
AdaSORS-II	&	92.86$\pm$0.01	&	99.64$\pm$0.01	&	162515.32$\pm$9431.53	&	&	67.10$\pm$0.01	&	98.20$\pm$0.01	&	687947.24$\pm$15972.15	\\
\hline
\end{tabular}
\end{small}
\end{center}
\vspace{-0.0in}
\end{table*}

Sector dataset is used for text categorization~\cite{mccallum1998comparison}. The scaled data consists of company web pages classified in a hierarchy of industry sectors. It is collected from web sites belonging to industry companies from various economic sectors. The whole dataset contains 9,619 entries. Each Sector entry is represented by 55,197 features.

For each dataset, data instances from each class are randomly split into training set (70\%) and test set (30\%). To generate a triplet $(\x_t,\x_t^+,\x_t^-)$, $\x_t$ is firstly randomly selected from the whole training set, then $\x_t^+$ is randomly selected from the subset of training set, which consists of the examples with the same class of $\x_t$, finally $\x_t^-$ is randomly selected from the rest of training set, which consists of the examples with different classes of $\x_t$.

To make a fair comparison, all algorithms adopt the same experimental setup. We use cross-validation to select the values of hyper parameters for all algorithms. Specifically, the parameters set by cross validation include: the $\mu$ parameter for LMNN ($\mu\in$\{0.1,0.2,...0.9\}), the $\gamma$ parameter for ITML ($\gamma\in$\{0.1,1,10\}), the regularization parameter $\eta$ for LEGO ($\eta\in$\{0.02,0.08,0.32\}), the $\lambda$ parameter for RDML ($\lambda\in$\{0.01,0.1\}), the scalar parameter $r$ for AROMA ($r\in$\{0.01,0.1,1,10,20...100\}) and the aggressiveness parameter $C$ for OASIS ($C\in$\{0.1,0.08,0.06,0.04...\}). All the evaluation results listed below are achieved by choosing  the parameters using cross validation.


\vspace{-0.05in}
\subsection{Evaluation Measures}
\vspace{-0.05in}
We evaluate the performance of the algorithms, by the sparsity of the model, the precision at top $k$ retrieval results, the mean average precision of retrieval results, and the running time.
\begin{itemize}
  \item The sparsity of a matrix $M$ is defined as Sparsity$= 1-\frac{\|M\|_0}{d^2}$, which is controlled by $\lambda$ in Eq.~\eqref{eqn:goal} and Eq.~\eqref{eqn:so-goal}.
  \item Precision at $top$-$k$  corresponds to the number of relevant results on the first $k$ query results.
  \item The \emph{mean average precision} (MAP) is used to evaluate the accuracy of the retrieval performance.
  \item The training time is also reported for each algorithm to show the computational efficiency.
\end{itemize}


\vspace{-0.05in}
\subsection{Performance Evaluation}
\vspace{-0.05in}
Table~\ref{tab:performance} summarizes the performance of all the compared similarity/metric learning algorithms over all the benchmark datasets, where results for LMNN on ``BBC" ``RCV1" and ``Sector" are not reported since the learning task can not be finished in 168 hours. From the results, we can draw several observations.

Firstly, the sparsity of Euclidean metric is very high, however its MAP performance  is much lower than those of other algorithms for most of cases, which implies the importance of metric learning for improving the information retrieval performance.

Secondly, among the baseline algorithms in comparison, we observe that the offline algorithm LMNN is generally significantly slower than the other online learning algorithms including ITML, LEGO, RDML, and OASIS, although it generally achieves better MAP than the online learning algorithms. This verifies the efficiency and scalability of online learning, which makes them more attractive for large-scale tasks and mining data streams.

Thirdly, among the online learning algorithms in comparison, we observe that the online similarity learning algorithms (OASIS and AROMA) generally achieve better or at least comparable MAP performance compared with online metric learning algorithms (ITML, LEGO, RDML), which indicates the effectiveness of similarity learning techniques, especially that learning similarity function without PSD constraint is effective. Compared with OASIS, AROMA generally achieves better MAP performance on all the datasets, which verifies the effectiveness of introducing second order information to improve the learning efficacy

Finally, comparing with all the baseline algorithms,  the proposed SORS algorithms (SORS, AdaSORS) achieve the highest test MAP performance on all the datasets. This shows that the proposed learning strategy is effective in improving the generalization ability of the learnt model. Secondly, by examining the sparsity of resulting matrix, we observe that SORS algorithms result in the sparsest model than the other baseline algorithms for most of the datasets (Although ITML produces sparser model for Gisette and RDML produces sparser models for Gisetter and BBC, their MAP performance is significantly worse). This implies that the proposed strategy can effectively improve the sparsity of the modal. In addition, compared with SORS-I and SORS-II, AdaSORS algorithms (AdaSORS-I and AdaSORS-II) generally achieve better or comparable MAP, and produce comparably or sparser models.

\begin{figure}[!htpb]
\centering
\begin{subfigure}
\centering
\includegraphics[width=2.55in]{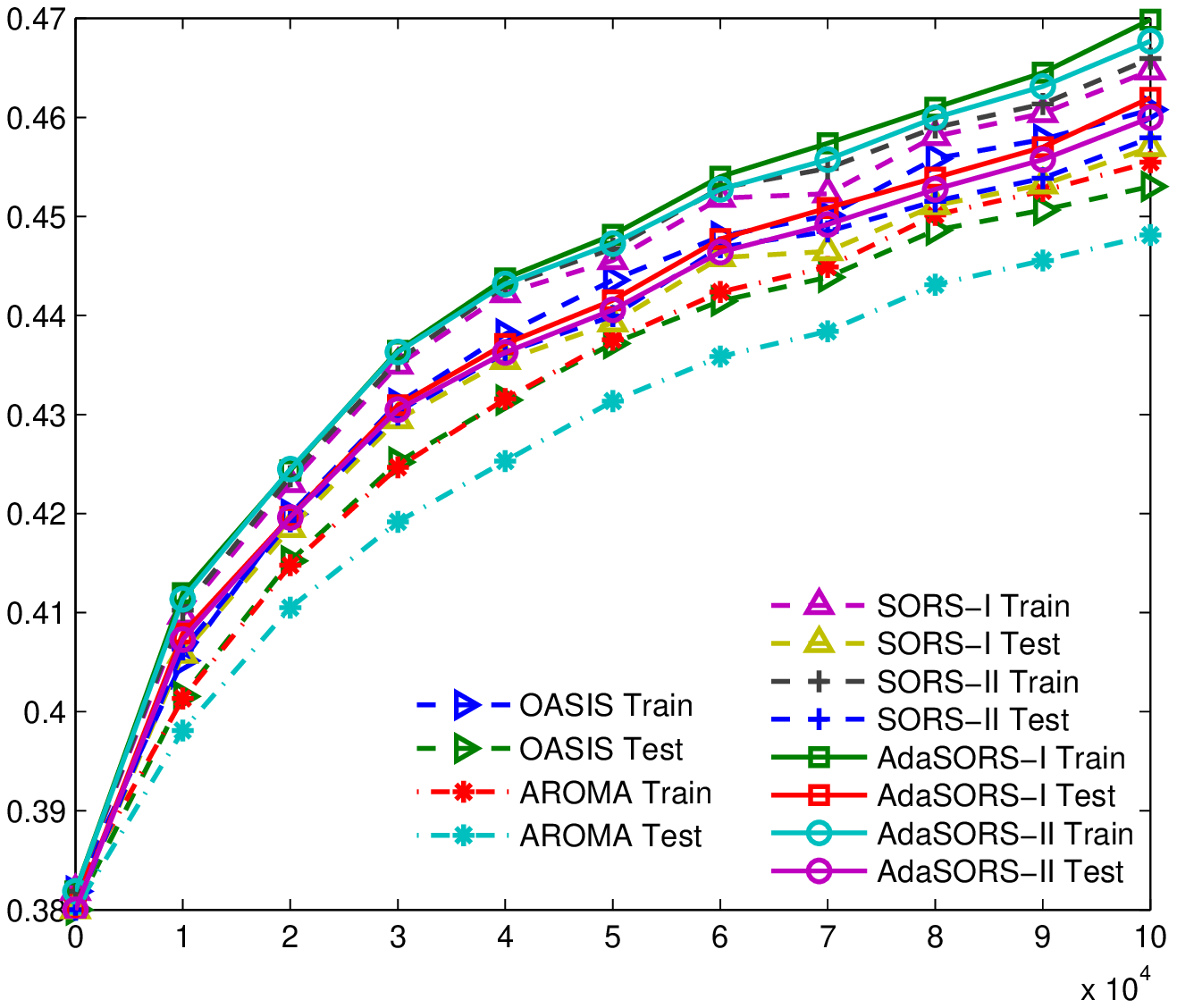}
\makebox[2.5in]{(a) Protein}
\end{subfigure}
\centering
\begin{subfigure}
\centering
\includegraphics[width=2.55in]{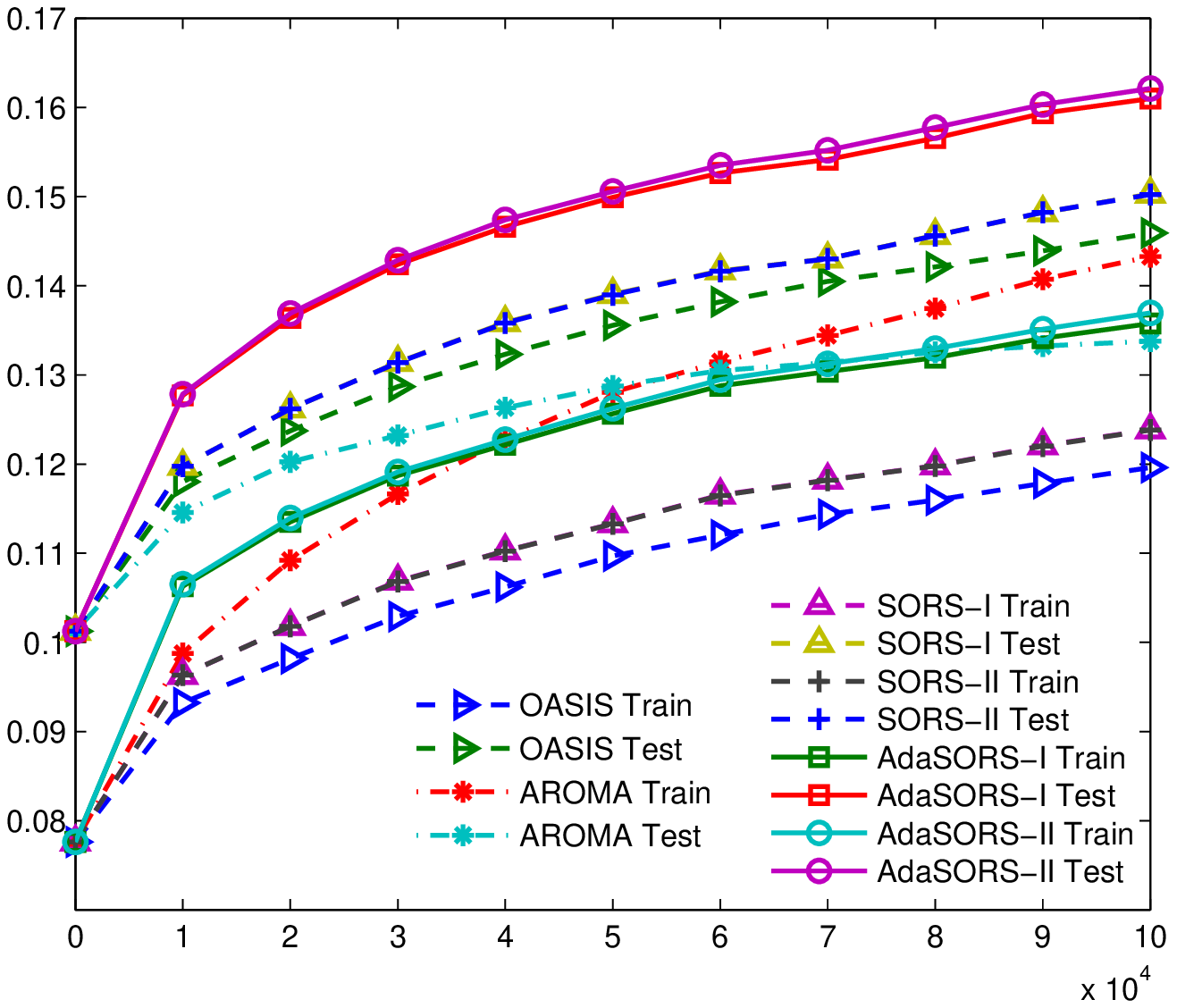}
\makebox[2.5in]{(b) Caltech50}
\end{subfigure}
\centering
\begin{subfigure}
\centering
\includegraphics[width=2.55in]{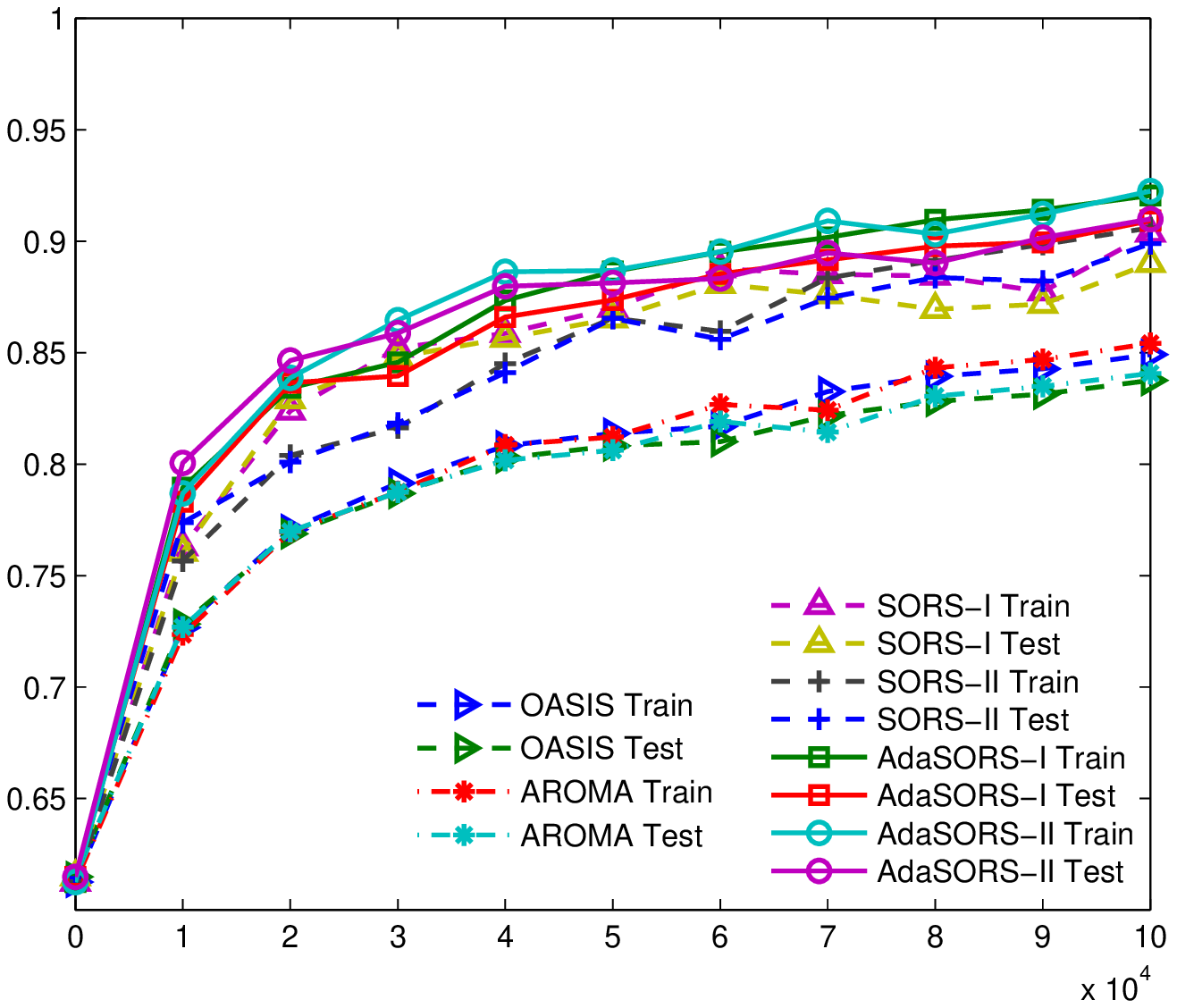}
\makebox[2.5in]{(c) Gisette}
\end{subfigure}
\centering
\begin{subfigure}
\centering
\includegraphics[width=2.55in]{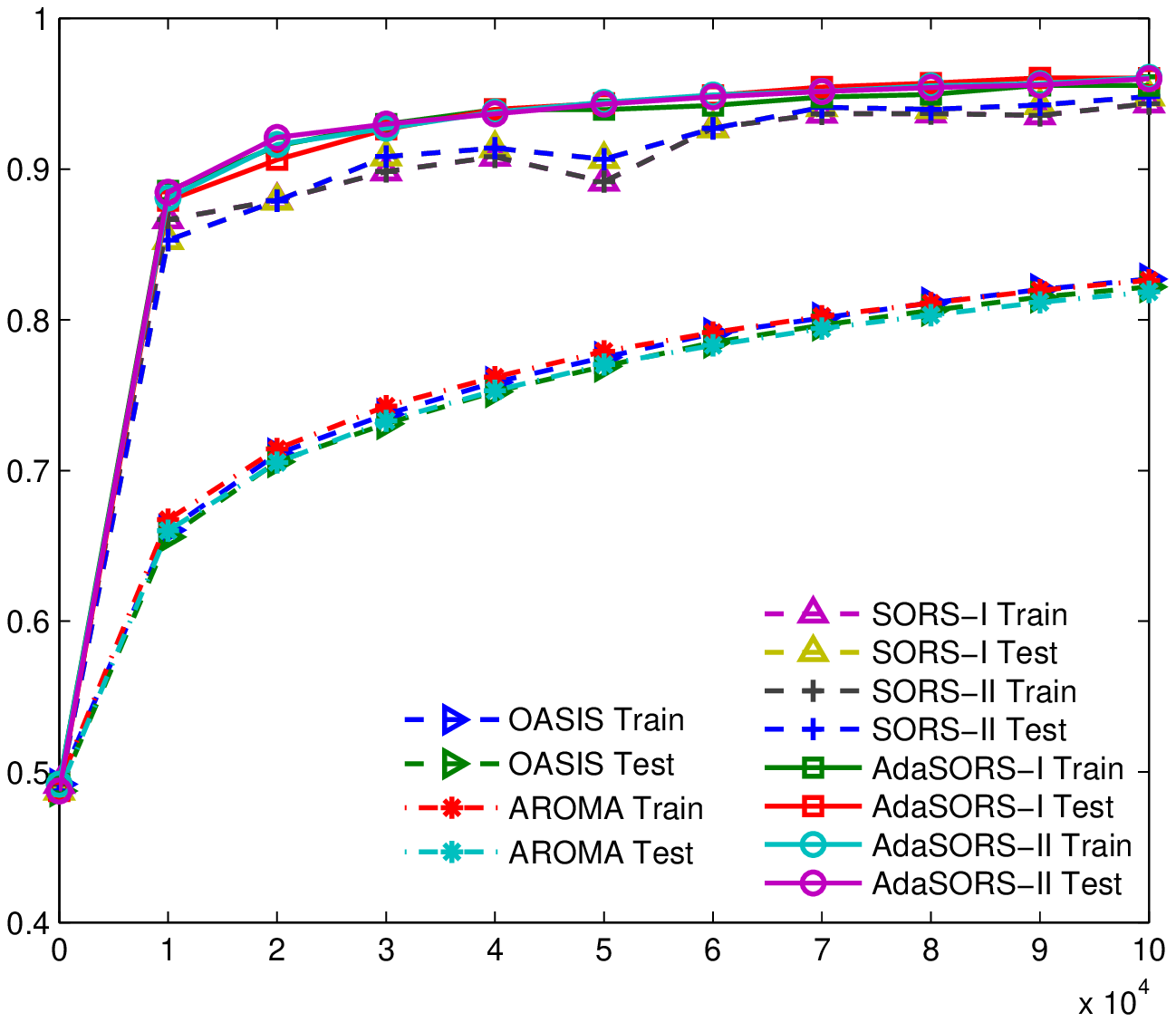}
\makebox[2.5in]{(d) BBC}
\end{subfigure}
\caption{MAP of online similarity learning algorithms as a function of the number of training iteration on four data sets.}\label{fig:MAP-iteration}
\end{figure}

Moreover, Figure~\ref{fig:MAP-iteration} shows the MAP performance of all the online similarity learning algorithms in comparison over trials. The parameters for these algorithms are selected by cross validation. Specifically, on the dataset ``Protein", the parameters are set as: $C=0.01$ for OASIS, $r=60$ for AROAM,  $\lambda=1.e-6$, $\eta=0.1$ for SORS-I \& SORS-II, and $\lambda=1.e-4$, $\eta=0.1$, $\delta=5$ for AdaSORS-I \& AdaSORS-II . On the dataset ``Caltech50'', the parameters are set as: $C=0.06$ for OASIS, $r=13$ for AROMA, $\lambda=1.e-6$, $\eta=0.1$ for  SORS-I \& SORS-II, and $\lambda=1.e-6$, $\eta=0.1$, $\delta=0.1$ for AdaSORS-I \& AdaSORS-II. On the dataset ``Gisette'', the parameters are set as $C=0.01$ for OASIS,  $r=90$ for AROMA, $\lambda=1.e-6$, $\eta=0.06$ for SORS-I \& SORS-II, and $\lambda=1.e-5$, $\eta=0.06$, $\delta=20$ for AdaSORS-I \& AdaSORS-II. On the dataset ``BBC'', the parameters are set as: $C=0.01$ for OASIS, $r=90$ for AROMA, $\lambda=1.e-6$, $\eta=0.1$ for SORS-I \& SORS-II, and $\lambda=1.e-6$, $\eta=0.1$, $\delta=0.1$ for AdaSORS-I \& AdaSORS-II. If not stated, the parameters for the rest of the figures are set as the same as here.

\begin{figure}[!htpb]
\centering
\begin{subfigure}
\centering
\includegraphics[width=2.65in]{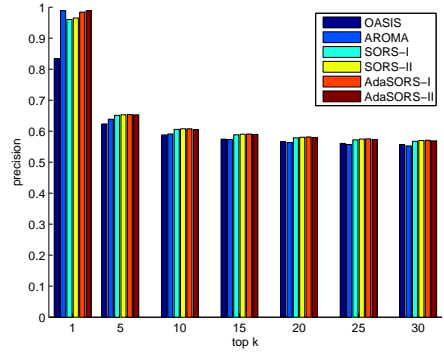}
\makebox[2.5in]{(a) Protein}
\end{subfigure}
\centering
\begin{subfigure}
\centering
\includegraphics[width=2.65in]{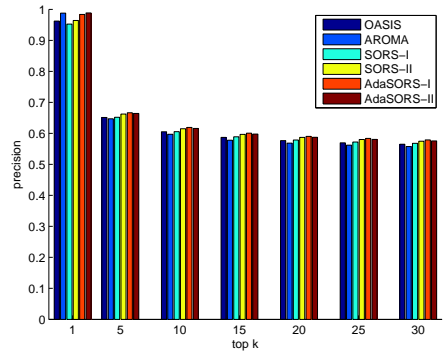}
\makebox[2.5in]{(b) Caltech50}
\end{subfigure}
\centering
\begin{subfigure}
\centering
\includegraphics[width=2.65in]{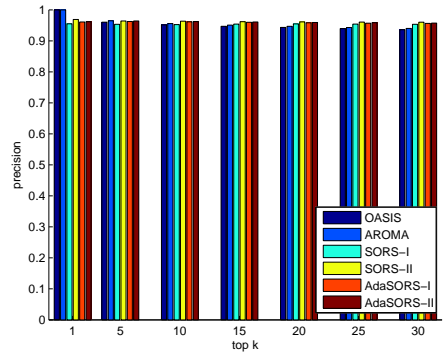}
\makebox[2.5in]{(c) Gisette}
\end{subfigure}
\centering
\begin{subfigure}
\centering
\includegraphics[width=2.65in]{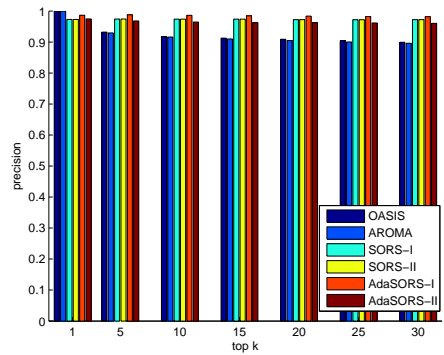}
\makebox[2.5in]{(d) BBC}
\end{subfigure}
\caption{Precision$@$k of the online similarity learning algorithms on four data sets.}\label{fig:Prec_k}
\end{figure}

\begin{figure}[!htpb]
\centering
\begin{subfigure}
\centering
\includegraphics[width=2.65in]{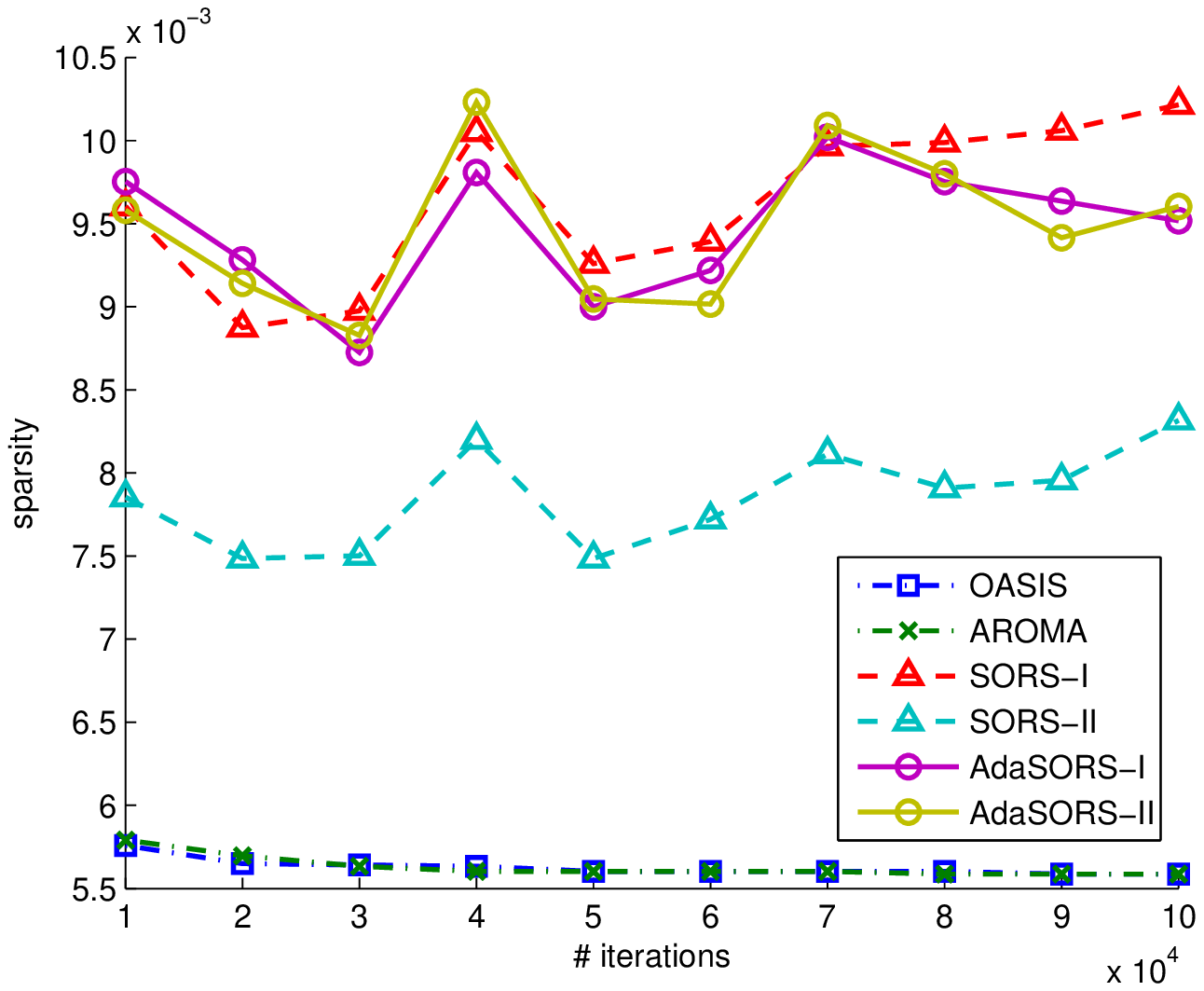}
\makebox[2.5in]{(a) Protein}
\end{subfigure}
\centering
\begin{subfigure}
\centering
\includegraphics[width=2.65in]{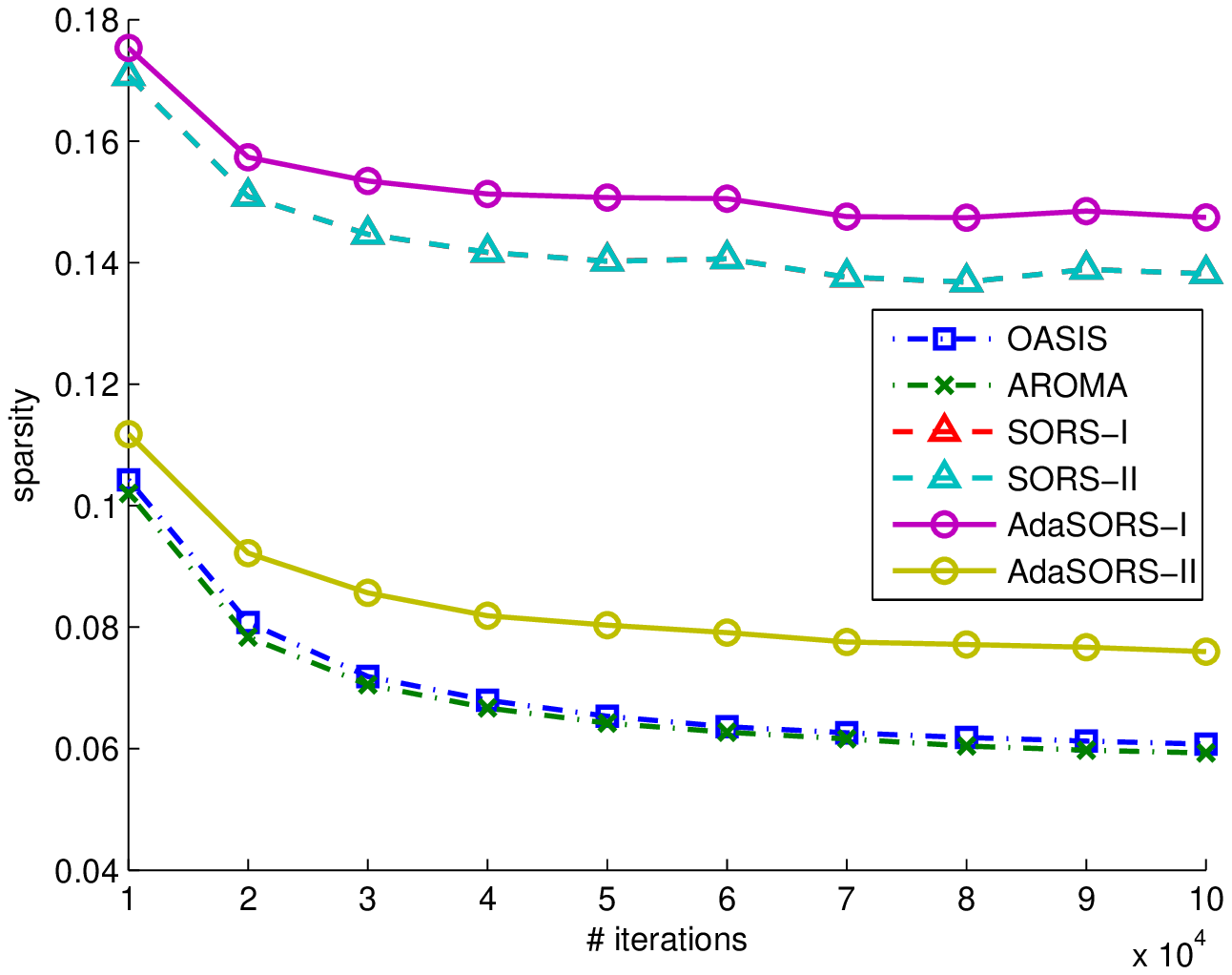}
\makebox[2.5in]{(b) Caltech50}
\end{subfigure}
\centering
\begin{subfigure}
\centering
\includegraphics[width=2.65in]{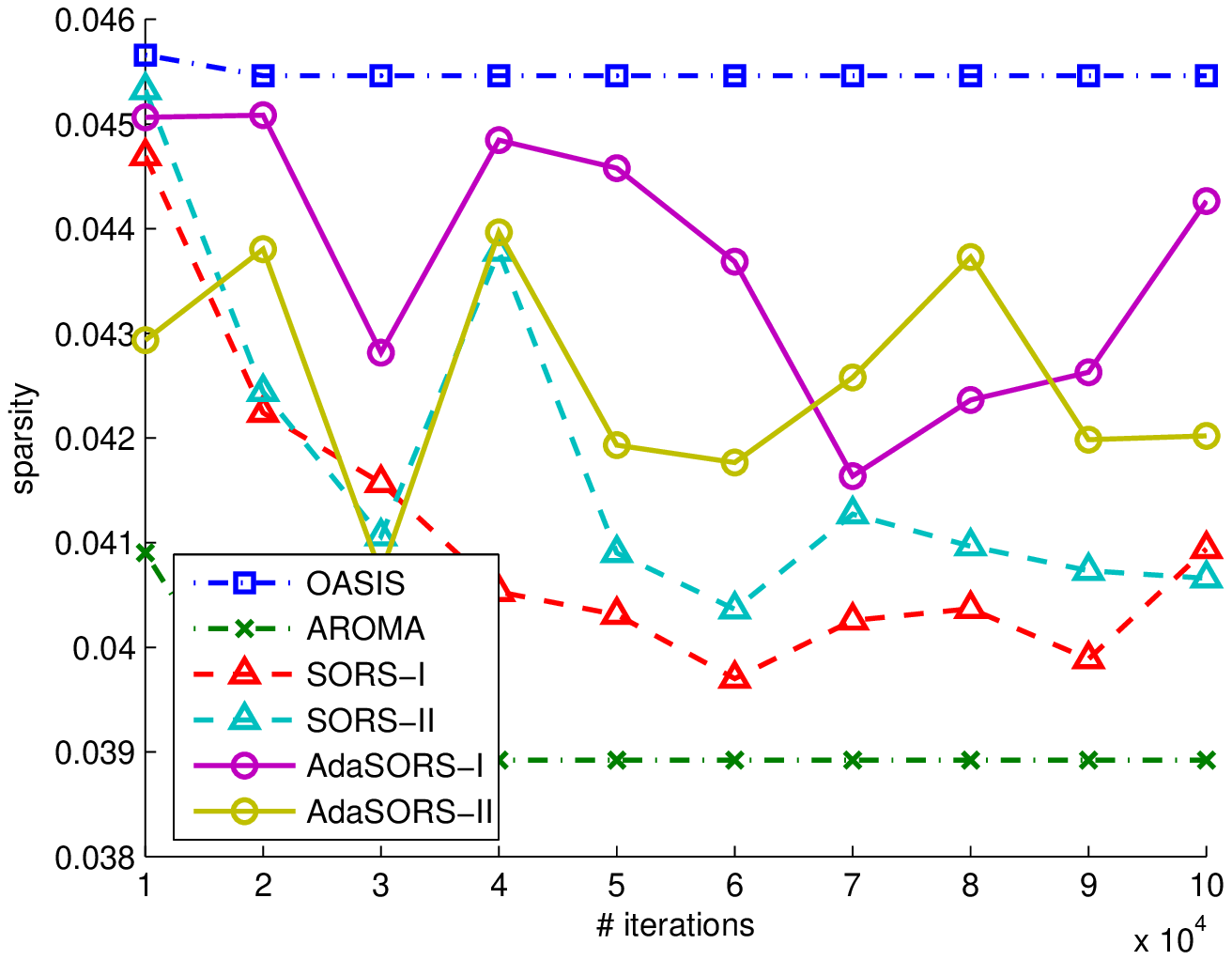}
\makebox[2.5in]{(c) Gisette}
\end{subfigure}
\centering
\begin{subfigure}
\centering
\includegraphics[width=2.65in]{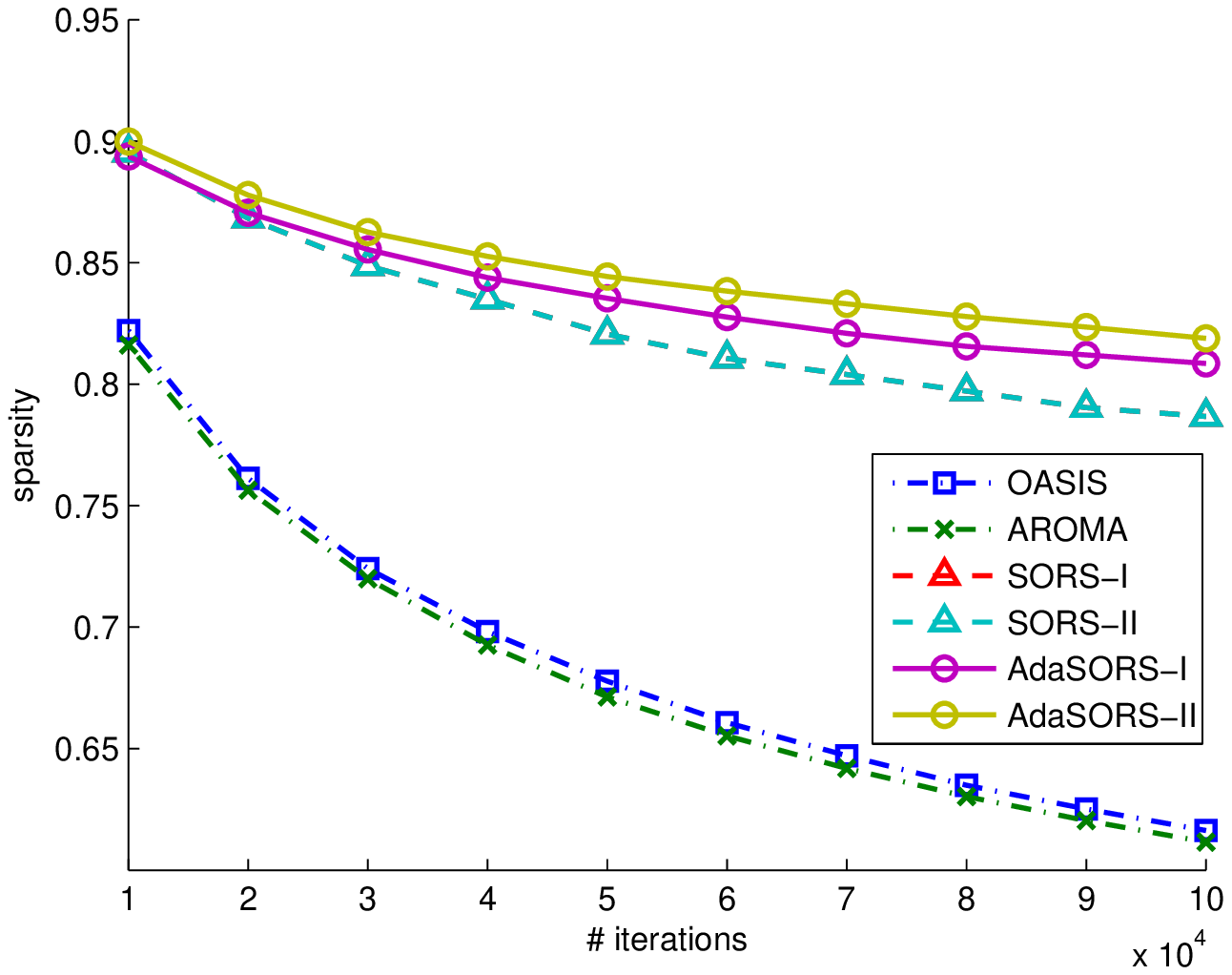}
\makebox[2.5in]{(d) BBC}
\end{subfigure}
\caption{Online sparsity of the online similarity learning algorithms on four data sets.}\label{fig:sparsity-iteration}
\end{figure}

From Figure~\ref{fig:MAP-iteration}, we can observe that generally SORS algorithms outperform OASIS and AROMA, which indicates the effectiveness of introduction of sparsity. In addition, AdaSORS algorithms generally can achieve the best MAP performance, which verifies the effectiveness of using second order information for improving learning efficacy. Overall, these observations are consistent with  Table~\ref{tab:performance}, which again verifies that the introduction of sparse regularization can efficiently remove noisy interactions between features and improve the generalization ability of the learning algorithms.

To further demonstrate the effectiveness of the proposed algorithms, Figure~\ref{fig:Prec_k} summarizes the precision$@k$  performance of the online similarity learning algorithms: OASIS, AORMA, SORS-I, SORS-II, AdaSORS-I and AdaSORS-II on four datasets. It can be  observed that AdaSORS and SORS algorithms overall outperform OASIS and AROMA on most of $k$ (except $k$=1). The reason for this is the fact SORS and AdaSORS algorithms are designed to optimize the AUC performance, which may not guarantee the best precision at top $k$. However, the proposed algorithms still achieve better precision at top $k$ for most of cases. All these observations verify the effectiveness of the proposed algorithms.

Moreover, we also estimate the online sparsity of the models produced by online similarity learning algorithms with respect to iterations on four datasets, illustrated in Figure~\ref{fig:sparsity-iteration}. We can observe the sparsity generally decreases as the number of iteration increases  for all the algorithms, which is due to the fact more and more rank-1 matrices are added to the model. In addition, AdaSORS and SORS algorithms generally achieve much higher sparsity than AROMA and OASIS (except for Gisette, where all the algorithms achieve comparable sparsity) and AdaSORS algorithms overall produce sparser models than SORS, which again verifies the effectiveness of the sparse regularization to reduce memory cost.

\vspace{-0.05in}
\subsection{Testing Efficiency}
\begin{figure}[!htpb]
\centering
\includegraphics[width=3.1in]{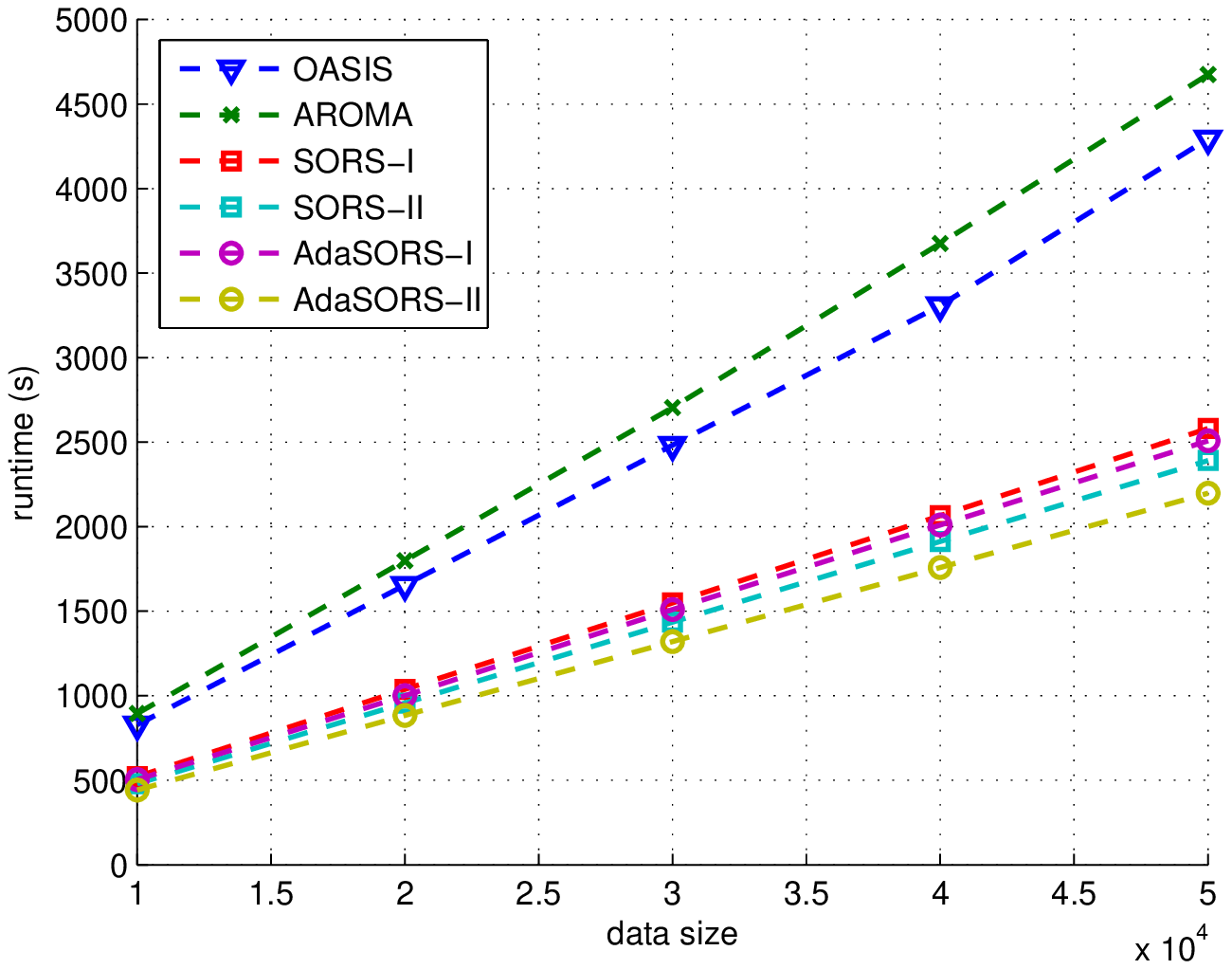}
\label{fig:bbc_runtime}
\caption{Comparison of the testing efficiencies of the models from 6 online similarity learning algorithms on ``BBC".}\label{fig:runtime}
\end{figure}

\vspace{-0.05in}
To evaluate the testing efficiency, Figure~\ref{fig:runtime} summarizes the query time of the models learnt using OASIS, AROMA, SORS-I, SORS-II, AdaSORS-I, and AdaSORS-II on ``BBC'' for test datasets with varied sizes, for which the sparsity of these models are already shown in the Table~\ref{tab:performance}.

We can observe that AdaSORS and SORS algorithms generally are more efficient than OASIS and AROMA over all the test sets of different sizes, since the models of these four algorithms are much sparser than OASIS and AROMA. In addition, when the number of queries is small, the time cost of non-sparse algorithms (OASIS, AROMA) and sparse algorithms (SORS-I, SORS-II, AdaSORS-I, AdaSROS-II) are not much different; however if the number of queries is very large, sparse algorithms generally are significantly efficient than non-sparse algorithms, e.g., AdaSORS-II method saves 50\% of test time when the size of data is 50K. This makes the proposed algorithms more attractive for large-scale high dimensional real-world applications.

%
%

\vspace{-0.05in}
\subsection{Sparsity-MAP Tradeoffs}
\begin{figure}[!htpb]
\vspace{-0.1in}
  \centering
  \includegraphics[width=3.1in]{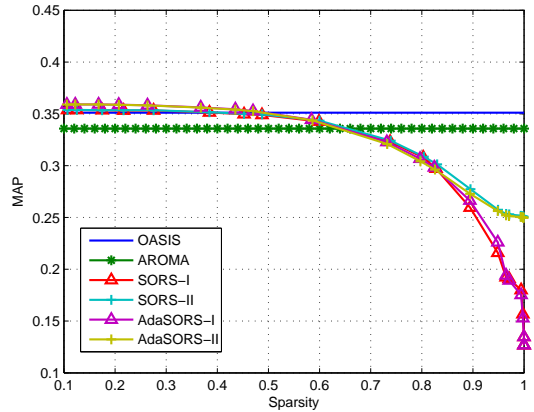}
  \caption{Tradeoff between MAP and Sparsity of models of online similarity learning algorithms on ``Caltech10'' dataset for 100K iterations.}\label{fig:caltech10_map_sparsity_sors}
\vspace{-0.1in}
\end{figure}

\vspace{-0.05in}
Finally, we evaluate the relationship between the MAP performance and the sparsity of the models of the online similarity learning algorithms on ``Caltech10'', which is demonstrated in Figure~\ref{fig:caltech10_map_sparsity_sors}, where the sparsity and MAP of OASIS are constant values.

We can observe that when the sparsity is small enough (less than 0.5), the MAPs performance of the proposed SORS and AdaSORS algorithms tend to be highest and stable, while when the sparsity is larger, increasing the sparsity generally leads to rapid deceasing of MAPs. However, the proposed SORS and AdaSORS algorithms allow us to choose a much sparser model, e.g. with around $15$ percentage sparsity, with  better or at least comparable MAP performance compared with OASIS and AROMA. In addition, when the sparsity is extremely high, SORS-II and AdaSORS-II can still achieve much better MAP performance than SORS-I and AdaSORS-I respectively, which indicates the importance of those diagonal elements for learning an effective similarity function.

\section{Conclusion}
\label{sec:con}
This paper presented a \emph{Sparse Online Relative Similarity Learning} (SORS) framework, and four scalable algorithms for similarity learning tasks with high dimensional features. This framework would like to learn a sparse model which can assign similar items with higher similar values.  We theoretically analyze the regret bounds of the proposed algorithms, and conduct a set of experiments by comparing with a number of competing algorithms. Promising empirical results shows that the proposed algorithms can effectively learn sparser model with better or at least comparable MAP performance.

\section*{Acknowledgment}
The work is partially supported by NSFC (No.61472149, No.71401128), the Fundamental Research Funds for the Central Universities (2015QN67), and the SRF for ROCS, SEM. The authors would like to thank Zhenzhong Lan for the fruitful comments on this work.




%
%
%

\bibliographystyle{IEEEtran}
\bibliography{icdm15}

\end{document}